\documentclass{article}
\usepackage{spconf,amsmath,epsfig}
\usepackage{amsmath}
\usepackage{amssymb}
\usepackage{xspace}
\usepackage{times}
\usepackage{epsfig}
\usepackage{graphicx}
\usepackage{amsmath}
\usepackage{amssymb}
\usepackage{algorithm}
\usepackage{algorithmic} 
\usepackage{setspace}
\usepackage{multirow}
\usepackage{booktabs}
\usepackage{bbm}

\usepackage{algorithm}
\usepackage{algorithmic}
\usepackage{booktabs}
\usepackage{subfigure}
\usepackage{amsmath}
\usepackage{cite}
\usepackage{float}
\usepackage{bbding}
\usepackage{amssymb}
\usepackage{newfloat}
\usepackage{listings}

\let\OLDthebibliography\thebibliography
\renewcommand\thebibliography[1]{
  \OLDthebibliography{#1}
  \setlength{\parskip}{0pt}
  \setlength{\itemsep}{0pt plus 0.3ex}
}

\begin{document}\sloppy
\topmargin=0mm

\def\x{{\mathbf x}}
\def\L{{\cal L}}

\title{Semantic-Assisted Image Compression}
%
\name{Qizheng Sun$^{\ast}$, Caili Guo$^{\ast\dagger}$, Yang Yang$^{\ast\dagger}$, Jiujiu Chen$^{\ast}$, Xijun Xue$^{\ast}$}
\address{$^{\ast}$Beijing Laboratory of Advanced Information Networks, \\
	Beijing University of Posts and Telecommunications, Beijing, China 100876 \\
	$^{\dagger}$Beijing Key Laboratory of Network System Architecture and Convergence, \\
	Beijing University of Posts and Telecommunications, Beijing, China 100876 \\
	Email: \{qizheng\_sun, guocaili, yangyang01, chenjiujiu\}@bupt.edu.cn, xuexj@chinatelecom.cn }

\maketitle

\begin{abstract}
Conventional image compression methods typically aim at pixel-level consistency while ignoring the performance of downstream AI tasks. 
To solve this problem, this paper proposes a Semantic-Assisted Image Compression method (SAIC), which can maintain semantic-level consistency to enable high performance of downstream AI tasks.
To this end, we train the compression network using semantic-level loss function. In particular, semantic-level loss is measured using gradient-based semantic weights mechanism (GSW). GSW directly consider downstream AI tasks' perceptual results. Then, this paper proposes a semantic-level distortion evaluation metric to quantify the amount of semantic information retained during the compression process. Experimental results show that the proposed SAIC method can retain more semantic-level information and achieve better performance of downstream AI tasks compared to the traditional deep learning-based method and the advanced perceptual method at the same compression ratio. 
\end{abstract}
\vspace{-0.2cm}
\section{Introduction}
\vspace{-0.15cm}
With the explosion of visual data on the internet, image compression has becoming a significant and fundamental task especially for caching and communication. There is a plethora of prior art on image compression including conventional methods such as JPEG and JPEG2000, and methods based on deep learning. Conventional methods compress images separately in transform, quantizer, and entropy code \cite{cheng2018deep}, which is inefficient. In contrast, deep learning-based compression methods are powerful due to the joint optimization of the entire compression model and excellent learning ability. Deep learning-based image compression have been explored by convolutional autoencoder (CAE) \cite{balle2016end}, recurrent network (RNN) \cite{lee2018context}, and generative adversarial networks (GAN) \cite{wu2020gan}. 

However, existing image compression methods aimed at maintaining pixel-level consistency. With the development of computer vision, a large number of compressed images need to be understood by downstream AI tasks such as image recognition, object detection, etc. Image compression should have both high visual quality and high performance of downstream AI tasks. Solely maintaining pixel-level consistency cannot guarantee the task performance.
\begin{figure}[t]
	\centering
	\includegraphics[width=0.8\linewidth]{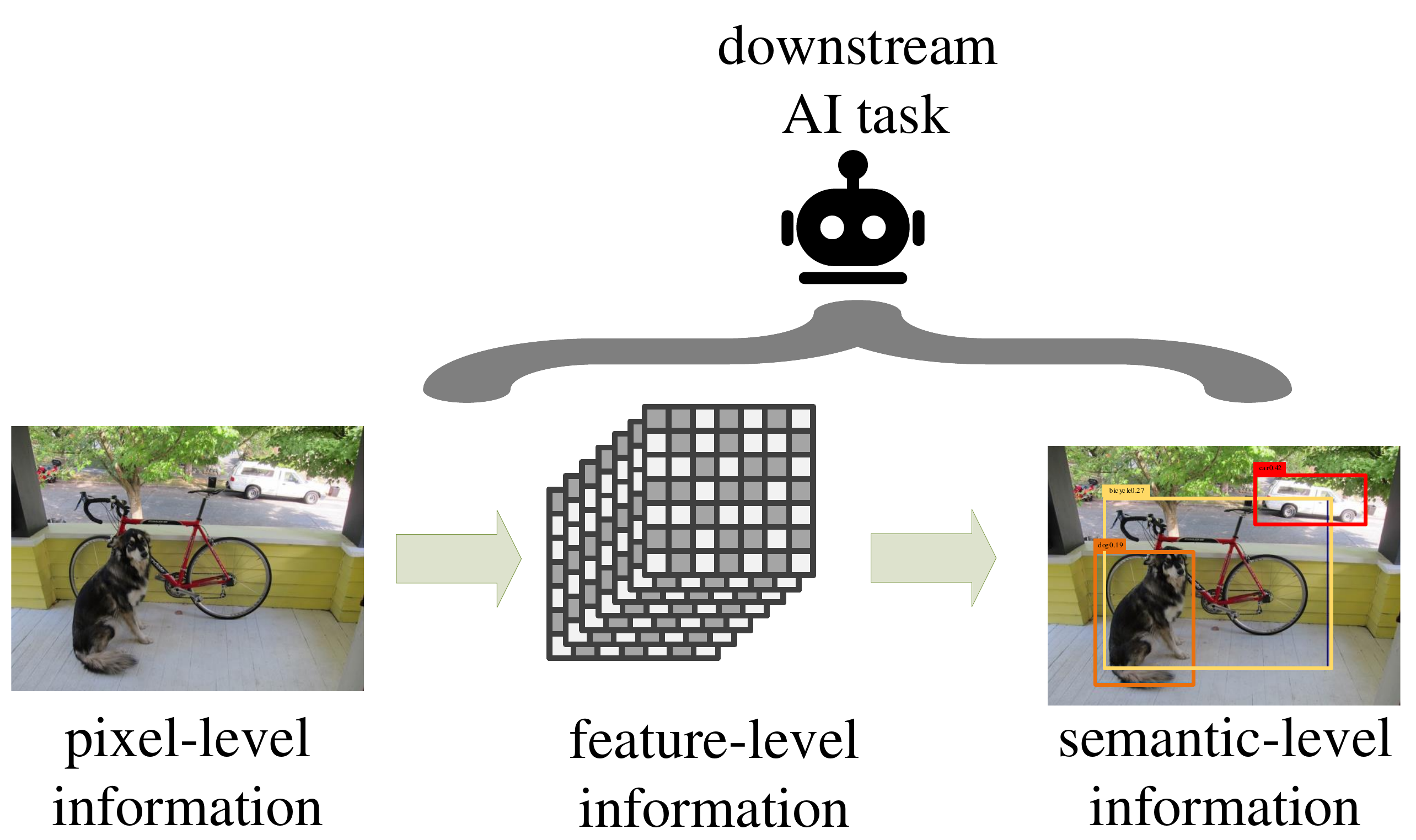}
	\caption{Images contain pixel-level information, and go through the downstream AI task, feature-level and semantic-level information can be obtained. Semantic-level information is the meaning finally understood by the downstream AI task, which is contained in perceptual results, such as locations and confidence scores.}    
	\vspace{-0.4cm}
	\label{introduction}
\end{figure}

To this end, Yang et al. \cite{yang2020discernible} proposed a discernible image compression method aiming at maintaining feature-level consistency of downstream AI tasks. Although features are sensitive, using feature maps as loss functions is still reliable, which is demonstrated through solid experiments in \cite{yang2020discernible}. To further illustrate the reliability, Table \ref{loss} shows that the feature-level distortion (mean square error of feature maps) decreases with the decrease of compression degree.

However, the performances of downstream AI tasks are ultimately determined by semantic-level information rather than feature-level information. This paper utilizes the fact that perceptual results contain all semantic-level information understood by the downstream AI task. Fig.\ref{introduction} illustrates the relationship among pixel-level, feature-level and semantic-level information. Although semantic-level information is further extracted from feature-level information and good feature-level information is more likely to extract good semantic-level information \cite{lecun2015deep}, they are essentially different. Feature-level information is the intermediate output of the downstream AI task, while semantic-level information is the final meaning understood by the downstream AI task. Thus, semantic-level information can influence AI task's performance directly. Taking object detection task as an example, the task performance mean average precision (mAP) is determined by the semantic-level information, which is contained in perceptual results (bounding boxes and confidence scores). Therefore, different from maintaining feature-level consistency, maintaining semantic-level consistency during the compression process has the potential to further improve the performance of downstream AI tasks.
\begin{table}
	\scriptsize
	\centering
	\caption{The mean square error of feature maps with different compression ratios using pixel-level and feature-level consistency as the loss function.}
	\label{loss}
	\setlength{\abovecaptionskip}{-0.2cm}
	\begin{tabular}{llllll}
		\toprule[1pt]
		compression ratio (BPP) & 0.125 bpp & 0.25 bpp & 0.5 bpp \\
		\hline
		pixel-level consistency &  2.7865 & 1.6392 & 1.4698 \\
		\hline
		feature-level consistency &  3.7512 & 2.0676 & 1.8129 \\
		\toprule[1pt]
	\end{tabular}
	\vspace{-0.5cm}
\end{table}

In this work, we propose a Semantic-Assisted Image Compression method (SAIC) that maintains semantic-level consistency during compression. The purpose is retaining semantic-level information of downstream AI tasks during compression and consequently obtain good task performance. The innovation is that utilizes semantic-level information of downstream AI tasks and closely combining the compression task with the downstream AI task. The main contributions of this work are:
\vspace{-0.2cm}
\begin{itemize}
	\item We propose a SAIC method, in which the semantic-level information of downstream tasks is utilized to assist image compression.
	\vspace{-0.2cm}
	\item We propose a gradient-based semantic weights mechanism (GSW) to obtain semantic-level importance, which directly considers downstream AI tasks' perceptual results.
	\vspace{-0.2cm}
	\item We propose a semantic mutual information metric (SI) to quantify the semantic-level distortion during compression process for specific downstream AI task.
\end{itemize}
\vspace{-0.45cm}
\section{Related Work}
\noindent\textbf{Image Compression.} \;There are several image compression methods based on deep learning, such as RNN-based networks  \cite{lee2018context}, CNN-based networks \cite{ma2019cnn} and generative adversarial networks (GAN) \cite{wu2020gan}, which use pixel-level difference as distortion and do not consider downstream tasks. In addition, some advanced works took content information \cite{li2018learning} and task information \cite{patwa2020semantic} \cite{yang2020discernible} into consideration. In particular, Li et al. \cite{li2018learning} considered edges and textures information without considering the task performance. Patwa et.al. \cite{patwa2020semantic} simultaneously accomplished classification and decoding using the same compact feature representation, which may hurt the task performance. Yang et al. \cite{yang2020discernible} utilized feature-level information in loss function. In conclusion, these works do not consider semantic-level information of the downstream AI task. The proposed SAIC focuses on retaining semantic-level information of downstream AI tasks during compression, and strive for satisfying task performance.

\noindent\textbf{Interpretable CNN.} We introduce interpretable CNN to extract semantic-level information. Selvaraju et al. \cite{selvaraju2017grad} proposed a Gradient-weighted Class Activation Mapping (Grad-CAM) method, which can make visual explanation for perceptual results. The factors that determine perceptual results inspire us to extract the useful semantic-level information. 
\vspace{-0.25cm}
\section{Method}
\begin{figure*}[t]
	\setlength{\abovecaptionskip}{-1cm}   
	\setlength{\belowcaptionskip}{-8cm}   
	\centering
	\includegraphics[width=0.9\linewidth]{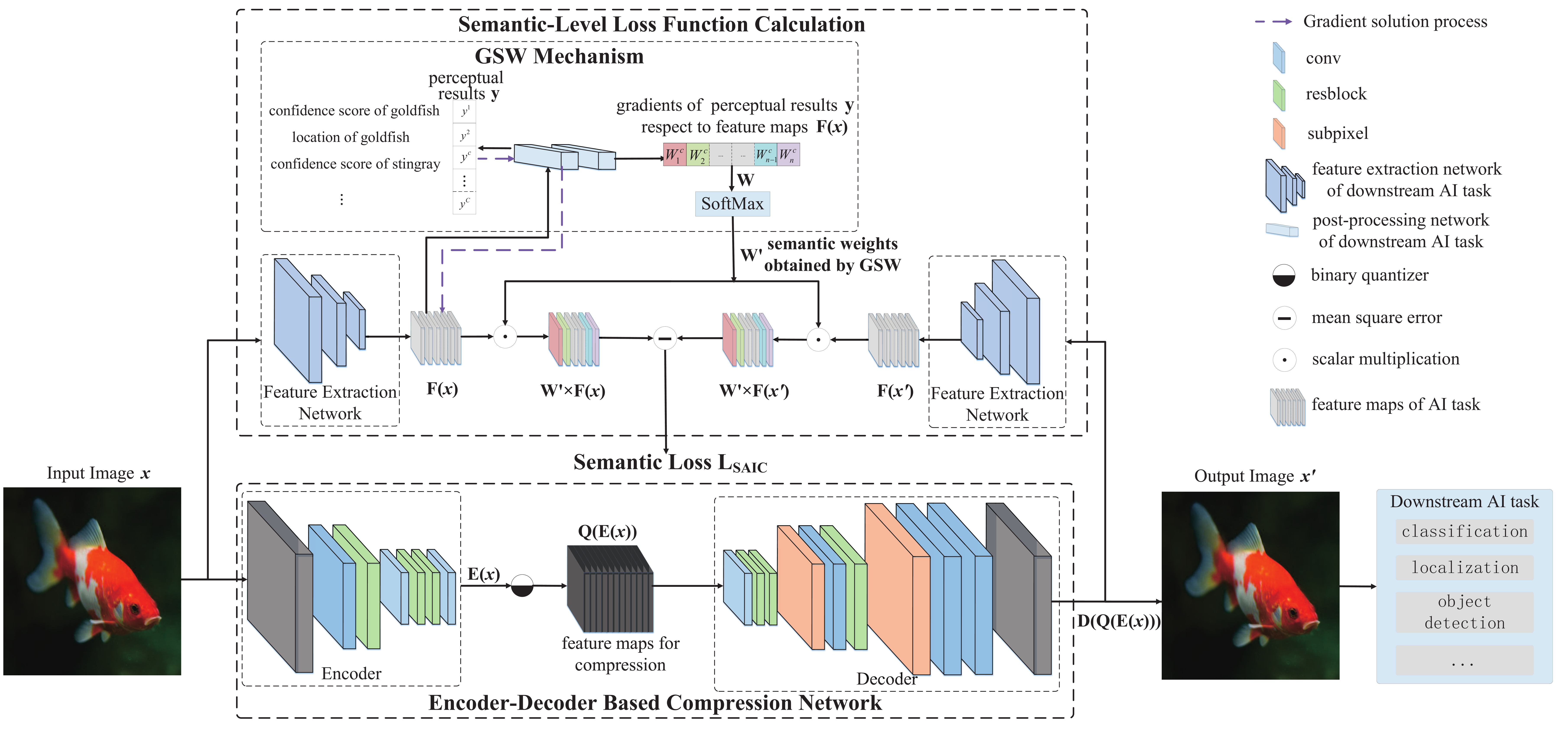}
	\vspace{-0.3cm}
	\caption{Architecture of the Semantic-Assisted Image Compression}
	\label{SAIC}
	\vspace{-0.4cm}
\end{figure*}
This section elaborates the proposed SAIC, which can retain semantic-level information of downstream AI tasks during compression process by using semantic-level loss function. As illustrated in Fig.\ref{SAIC}, the proposed SAIC architecture is composed of two modules, i.e. semantic-level loss function calculation module and encoder-decoder based compression network module. In particular, we first utilize semantic-level loss function calculation module to calculate the semantic-level loss function $\rm{L_{SAIC}}$. Then we use the semantic-level loss function $\rm{L_{SAIC}}$ to train the encoder-decoder based compression network module. 

As shown in Fig.\ref{SAIC}, the original image ${\boldsymbol{x}}$ and the compressed image ${\boldsymbol{x}}'$ first pass a pre-trained CNN to obtain the feature maps ${\boldsymbol{F}}({\boldsymbol{x}})$ and ${\boldsymbol{F}}({\boldsymbol{x}}')$. Then, we calculate the semantic-level weights ${\bf{W'}}$ by GSW mechanism using the gradient from perceptual results to feature maps. Next, we use ${\boldsymbol{F}}({\boldsymbol{x}})$, ${\boldsymbol{F}}({\boldsymbol{x}}')$ and ${\bf{W'}}$ to compute the semantic-level loss function ${\rm L_{SAIC}}$. Finally, we utilize the semantic-level loss function ${\rm L_{SAIC}}$ to train the encoder-decoder based compression network.
\vspace{-0.3cm}
\subsection{GSW Mechanism}
\vspace{-1ex}
GSW mechanism can obtain semantic weights, which is used to construct SAIC method. Semantic-level information is contained in perceptual results, and task performance is determined by perceptual results. GSW mechanism uses the gradients of perceptual results respect to feature maps, which can naturally represent the contribution of feature maps respect to perceptual results. 

We first pre-train a feature extraction network of downstream AI task to provide input for GSW mechanism, which can be expressed as:
\vspace{-0.1cm}
\begin{eqnarray}
	\rm {F}({\theta _1},{\boldsymbol{x}}) = {\boldsymbol{F}} = \left\{ {{{\boldsymbol{f}}_1},{{\boldsymbol{f}}_2},...,{{\boldsymbol{f}}_K}} \right\} \in {{\rm{{\cal R}}}^{K \times M \times N}},
\end{eqnarray}
where $ {\boldsymbol {f}_k} (k \in \left\{ {1,2,...,K} \right\})$ is the $k$-th feature map, ${\theta _1}$ is the fixed parameter of feature extraction network, and $M$, $N$ and $K$ represent the width, height, and total number of the feature maps, respectively. Then, semantic-level information is further extracted from feature-level information by post-processing network. In this process, perceptual results ${{\boldsymbol{y}}} = [y^1, y^2,...,y^c,...,y^C]$ is obtained, where $c \in \left\{ {1,...,C} \right\}$. For example, the perceptual results ${\boldsymbol{y}}$ are confidence scores in classification task, while confidence scores and localization information in object detection task. Obviously, as the input of semantic extraction network, feature maps ${\boldsymbol{F}}$ have different semantic-level importance degrees to obtain the semantic-level perceptual results for downstream AI task.

Then, the gradients of perceptual results respect to feature maps are used to quantify the semantic-level importance degrees, which is a matter of course due to the meaning of gradient.
We compute the gradient of $c$-th perceptual result ${{y}^c}$ with respect to the $k$-th feature map ${{\boldsymbol {f}}_k}$, i.e.$\frac{{\partial {{y}^c}}}{{\partial {{\boldsymbol{f}}_k}}}$. Then use global-average-pooled over the width $M$ and height $N$ dimensions (indexed by $m$ and $n$ respectively) to obtain the importance weights  $w_k^c$ by:
\vspace{-0.15cm}
\begin{eqnarray}
	w_k^c = \frac{1}{{M \times N}}\sum\limits_{m = 1}^M {\sum\limits_{n = 1}^N {\frac{{\partial {{y}^c}}}{{\partial {\boldsymbol{f}}_k^{}}}} },
\end{eqnarray}
where ${{\boldsymbol{f}}_k} \in {{\rm{\bf{{\cal R}}}}^{M \times N}}$. 
To obtain the semantic weights of the whole perceptual results $ {{\boldsymbol{y}}} $, we compute the average value:
\vspace{-0.15cm}
\begin{eqnarray}
	w_k^{} = \frac{1}{C}\sum\limits_{c = 1}^C{w_k^c}.
\end{eqnarray}

Thus, we obtain channel-wise semantic weights ${\bf{W = }}\left\{ {{w_1},{w_2},...,{w_K}} \right\} \in {{\rm{{\cal R}}}^K}$, where $w_k^{} (k \in \left\{ {1,2,...,K} \right\})$ represents the semantic importance degree of $k$-th channel’s feature map. The channel-wise semantic weights can be obtained under any dimension of perceptual results.

However, the value of ${\bf{W}}$ is too small to be directly used in the loss function, since it may cause the slow convergence. Therefore, we utilize parameter $\tau$, and map the weights ${\bf{W}}$ to the weights ${\bf{W}}'$ by:
\begin{eqnarray}
	{\bf{W}}' = r \times \rm{SoftMax} (\tau \times {\bf{W}}),
\end{eqnarray}
where $\tau$ is a temperature hyper-parameter, and $r$ is a constant. ${\bf{W}}'$ is appropriate for semantic-level loss function $\rm{L_{SAIC}}$. 
The temperature hyper-parameter $\tau$ can control the tightness of the semantic weights' distribution. Finally, we multiply a constant $r$ to make the final semantic-level loss value at a reasonable magnitude, which will not affect gradient updates. Note that due to GSW mechanism, semantic weights are task-specific, so SAIC is task-specific.
\vspace{-1.5ex}
\subsection{Semantic-level Loss Function calculation}
\vspace{-1ex}
We introduce this section to formulate semantic-level loss function $\rm{L_{SAIC}}$ for the image compression network. Generally, the loss function of traditional deep learning-based image compression network of encoder-decoder structure can be written as:
\vspace{-0.25cm}
\begin{eqnarray}
	\mathop {\min }\limits_{{\theta _2},{\theta _3}} \frac{1}{B}\sum\limits_{b = 1}^B {{{\left\| {\rm {D}({\theta _3},\rm {Q}(\rm {E}({\theta _2},{{\boldsymbol{x}}_b}))) - {{\boldsymbol{x}}_b}} \right\|}^2}},
\end{eqnarray}
where $B$ (batch size) is the number of images per iteration, $b$ is the index of images, ${{\boldsymbol{x}}_b}$ is $b$-th image. $\rm{E}( \cdot )$ is the encoder network with parameter ${\theta _2}$ for compressing the given image ${{\boldsymbol{x}}_b}$. $\rm {Q}( \cdot )$ is the quantizer. $\rm{D}( \cdot )$ is the decoder network with parameter ${\theta _3}$ for recovering the compressed latent features to images ${{\boldsymbol{x}}_b}'$. The compressed image can be written as:
\begin{eqnarray}
	\vspace{-0.2cm}
	{{\boldsymbol{x}}_b}' = \rm{D}({\theta _3},\rm{Q(E}({\theta _2},{{\boldsymbol{x}}_b}))).
\end{eqnarray}
For the sake of convenience, we use ${\boldsymbol{e}} = \rm{E}({\boldsymbol{x}}),{\boldsymbol{q}} = \rm{Q}(E({\boldsymbol{x}})),{\boldsymbol{d}} = \rm{D(B(E}({\boldsymbol{x}})))$ to represent the output of encoder, quantizer and decoder, respectively. Note that we use binary quantization to map the encoder output ${\boldsymbol{e}}$ to 0 or 1, which can be expressed as:
\vspace{-0.15cm}
\begin{eqnarray}
	\rm{Q}({\boldsymbol{e}}) = \left\{ {\begin{array}{*{20}{c}}
			{1,}&{{\boldsymbol{e}} > 0.5,}\\
			{0,}&{{\boldsymbol{e}} \le 0.5.}
	\end{array}} \right.
\end{eqnarray}

The pre-trained network extract feature-level information from original images and compressed images respectively by:
\begin{eqnarray}
	{\rm{F}({\theta _1},{{\boldsymbol{x}}_b}) = {\boldsymbol{F}}_b} = \left\{ {{\boldsymbol{f}}_1^b,{\boldsymbol{f}}_2^b,...,{\boldsymbol{f}}_K^b} \right\},
\end{eqnarray}
\vspace{-0.7cm}
\begin{eqnarray}
	{\rm{F}({\theta _1},{{\boldsymbol{x}}_b}') = {\boldsymbol{F}}_b}' = \left\{ {{\boldsymbol{f}}_1^{b'},{\boldsymbol{f}}_2^{b'},...,{\boldsymbol{f}}_K^{b'}} \right\},
\end{eqnarray}
where $k$-th feature map of original image $b$ is ${\boldsymbol{f}}_k^b$ and $k$-th feature map of compressed image $b'$ is ${\boldsymbol{f}}_k^{b'}$, $k \in \left\{ {1,2,...,K} \right\}$. 
Then, we utilize the channel-wise $K$-dimension semantic weights ${\bf{W}}'{\bf{ = }}\left\{ {{w_1}',{w_2}',...,{w_K}'} \right\} \in {{\rm{{\cal R}}}^K}$ obtained by GSW to weight image's feature maps as image's semantic-level information. Therefore, the loss function of SAIC can be written as:
\begin{eqnarray}\label{eq10}
	\begin{array}{l}
		\rm{{L_{SAIC}}}({\theta _2},{\theta _3}) = \\
		\frac{1}{B}\sum\limits_{b = 1}^B {\sum\limits_{k = 1}^K { { {{w_k}' \times {{\left\| {{\boldsymbol{f}}_k^{b'} - {\boldsymbol{f}}_k^b} \right\|}^2}} } } } .
	\end{array}
\end{eqnarray}
We summarize the steps of SAIC as Algorithm \ref{algorithm1}. In step 1, we pre-train the downstream AI network. In step 2, we obtain the semantic weights. In step 3-15, we compute $\rm{{L_{SAIC}}}({\theta _2},{\theta _3})$ and use it to train the compression network.
\begin{algorithm}[t]
	\setlength{\abovedisplayskip}{3cm}
	\caption{SAIC method}
	\label{algorithm1}
	\textbf{Input}: An image dataset $\left\{{{\boldsymbol{x}}^1},...,{{\boldsymbol{x}}^n}\right\}$ with $n$ images.\\
	\textbf{Parameter}: Encoder parameter ${\theta _2}$, decoder parameter ${\theta _3}$.\\
	\textbf{Output}: Parameters ${\theta _2}$ and ${\theta _3}$, compressed images, hidden feature maps ${\boldsymbol{q}}$.
	\begin{algorithmic}[1] 
		\STATE Pre-train the downstream network with parameter ${\theta _1}$, and fixed ${\theta _1}$ in the following operation.
		\STATE Obtain the semantic weights ${{\bf{W}}'}$ using GSW.
		\STATE Initialize compression network parameters ${\theta _2}$ and ${\theta _3}$.
		\WHILE{ not converged }
		\STATE Randomly select a batch of b images ${{\boldsymbol{x}}^1},...,{{\boldsymbol{x}}^b}$
		\FOR{ $i$=1 to $b$ }
		\STATE Compress the given image ${\boldsymbol{x}}^i$, ${{\boldsymbol{e}}^i} \leftarrow E({{\boldsymbol{x}}^i})$
		\STATE Quantize	${{\boldsymbol{e}}^i}$, ${{\boldsymbol{q}}^i} \leftarrow Q({{\boldsymbol{e}}^i})$
		\STATE Decode the data ${{\boldsymbol{x}}^{i'}} \leftarrow \rm{D}({{\boldsymbol{q}}^i})$
		\STATE Extract the features ${{\boldsymbol{F}}_i} \leftarrow \rm{F}({{\boldsymbol{x}}^i})$ and ${{\boldsymbol{F}}_i}'\leftarrow \rm{F}({{\boldsymbol{x}}^{i'}})$
		\ENDFOR
		\STATE Calculate ${\rm{L_{SAIC}}}({\theta _2},{\theta _3})$ according to Eq.\eqref{eq10}
		\STATE Update ${\theta _2}$ and ${\theta _3}$ according to ${\rm{L_{SAIC}}}({\theta _2},{\theta _3})$
		\ENDWHILE
		\STATE \textbf{return} The optimal compression model
	\end{algorithmic}
\end{algorithm}
\vspace{-0.3cm}
\subsection{SI Estimation}
\vspace{-1ex}
To quantify the semantic-level distortion during compression process for specific downstream AI task, we propose a semantic mutual information metric (SI). SI is the mutual information of all perceptual results, which contain all semantic-level information of the downstream AI task.
Let the perceptual results of the original image ${{\boldsymbol{x}}_b}$ be ${\boldsymbol{y}}_b$, and the perceptual results of compressed image ${{\boldsymbol{x}}_b}'$ be ${{\boldsymbol{y}}_b}'$.
However, it is challenging to estimate SI between ${\boldsymbol{y}}_b$ and ${{\boldsymbol{y}}_b}'$, since the entropy of the original image dataset is mathematical intractable. 
We utilize CLUB \cite{cheng2020club} to estimate SI due to its excellent accuracy. In particular, we first input ${\boldsymbol{y}}_b$ and ${{\boldsymbol{y}}_b}'$ to train a SI estimation network, from which we can obtain the mean and variance of ${{\boldsymbol{y}}_b}'$. Then we can compute the conditional probability ${p({\boldsymbol{y}}_b}'\mid {\boldsymbol{y}}_b)$ using the mean and variance. Finally, we compute SI using ${p({\boldsymbol{y}}_b}'\mid {\boldsymbol{y}}_b)$ by:
\vspace{-0.1cm}
\begin{eqnarray}
	\begin{array}{*{20}{c}}
		{{{\rm{I}}_{{\rm{CLUB}}}}({\boldsymbol{y}}_b;{{\boldsymbol{y}}_b}'):=
			\mathbbm{E}{_{p({\boldsymbol{y}}_b,{{\boldsymbol{y}}_b}')}}[\log p({{\boldsymbol{y}}_b}'\left| {{\boldsymbol{y}}_b} \right.)]}\\
		{ - \mathbbm{E}{_{p({\boldsymbol{y}}_b)}}\mathbbm{E}{_{p({{\boldsymbol{y}}_b}')}}[\log p({{\boldsymbol{y}}_b}'\left| {{\boldsymbol{y}}_b} \right.)]}
	\end{array}.
\end{eqnarray}
\vspace{-0.6cm}
\section{Experiment}
\begin{figure*}[htbp]
	\centering
	\setlength{\abovecaptionskip}{-0.5cm}
	\subfigure[Original]{
		\begin{minipage}[t]{0.18\linewidth}
			\centering
			\includegraphics[width=1.2in]{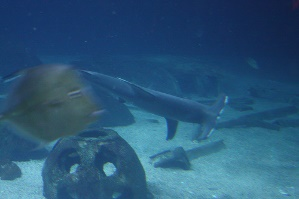}
		\end{minipage}%
	}
	\subfigure[TDIC:Electric ray \XSolidBrush
	]{
		\begin{minipage}[t]{0.18\linewidth}
			\centering
			\includegraphics[width=1.2in]{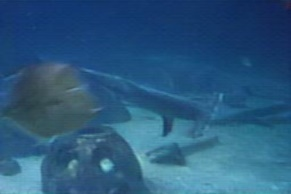}
		\end{minipage}%
	}%
	\subfigure[APIC:stingray \XSolidBrush
	]{
		\begin{minipage}[t]{0.18\linewidth}
			\centering
			\includegraphics[width=1.2in]{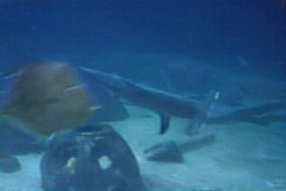}
		\end{minipage}
	}%
	\subfigure[SAIC:hammerhead \CheckmarkBold]{
		\begin{minipage}[t]{0.17\linewidth}
			\centering
			\includegraphics[width=1.2in]{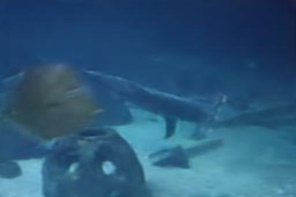}
		\end{minipage}
	}
	\centering
	\vspace{-0.2cm}
	\caption{Recognition results of compressed images using a pre-trained ResNet on an image labeled as hammerhead. From left to right are the original image and compressed images using TDIC, APIC and SAIC, respectively. Recognition labels are shown at the bottom of images. There are little visual difference, while large difference in perceptual results for downstream AI task.}
	\label{fish}
	\vspace{-0.4cm}
\end{figure*}
\vspace{-2.5ex}
\subsection{Dataset}
\vspace{-1ex}
For classification task, we carry out experiments on STL and ImageNet datasets. STL dataset \cite{coates2011selecting} has 10 classes, and each class contains 1300 images in PNG format. In terms of ImageNet \cite{deng2009imagenet}, we use about 9000 images in JPEG format. For object detection task, the experiments adopt Pascal VOC dataset \cite{everingham2010pascal}, which contains 27088 images in JPG format.
\vspace{-2.5ex}
\subsection{Evaluation Metrics}
\vspace{-1ex}
We use semantic mutual information (SI), accuracy (ACC), F1-score and mean average precision (mAP) to evaluate the semantic-level distortion. We use peak signal to noise ratio (PSNR), structural similarity index measure (SSIM) and mean square error per pixel (MSE) to evaluate the pixel-level distortion. We use compression ratio to evaluate the compression degree.
\vspace{-3ex}
\subsection{Comparison Methods}
\vspace{-1ex}
We use two baseline schemes for comparison: traditional deep learning-based image compression method (TDIC) and advanced perceptual image compression method (APIC). Note that the proposed SAIC and the two baseline schemes have different loss functions. For fair comparison, we use the same experiment conditions except for the loss function, including pre-trained downstream AI task network, the encoder-decoder based compression network structure and so on. TDIC, APIC and SAIC use pixel-level loss function, feature-level loss function \cite{yang2020discernible} and semantic-level loss function, respectively. 
\vspace{-2.5ex}
\subsection{Implementation Details}
\vspace{-1ex}
To reduce training costs and promote extensibility, we use the two-stage approach to train the APIC and SAIC model. In the first stage, the TDIC model is pre-trained for 3-4$\times {10^5}$ steps on different datasets with a batch size of 32 and a learning rate of 1$\times {10^{-5}}$. In the second stage, we finetune using APIC and SAIC methods for 2-3$\times {10^4}$ steps with a batch size of 32 and a learning rate of 1$\times {10^{-5}}$. For fair comparisons, to let the data load sequence be random and consistent, we fix random seed. Thus, the experiment is stable and repeatable. See appendix for specific experimental conditions.
\vspace{-2.5ex}
\subsection{Experiments on Classification Task}
\vspace{-1ex}
Table \ref{table1}, Table \ref{table2} and Table \ref{table3} show classification results. It is apparent that the proposed SAIC can always obtain better SI, ACC and F1-score in different datasets, different compression ratios and different classification network structures. As we can observe, SI, ACC and F1-score have the same trend, and ACC is very close to F1-score. SAIC can retain more useful semantic-level information in the compressed image due to the design of semantic-level loss function, and deservedly can obtain competitive SI, ACC and F1-score value. In addition, the conventional compression metrics PSNR, MSE and SSIM are as good as APIC method. These results suggest that without damaging the traditional indicators, we significantly reduce the semantic-level distortion and promote the ACC and F1-score of the downstream AI task.

As shown in Fig.\ref{fish}, the four sub-figures have little visual difference, while have totally different perceptual results for downstream AI task. See appendix for more examples. Some minor distortions caused by compression can lead to error perceptual results in downstream AI task. TDIC and APIC make a mistake since pixel-level and feature-level consistency cannot guaranty perceptual results. In contrast, the proposed SAIC directly focus on semantic-level consistency, thus can make right perceptual results and has the potential to own a better task performance.
\begin{table}
	\scriptsize
	\vspace{-0.3cm}
	\centering
	\caption{Classification results on the STL dataset with 0.125 bpp using ResNet18 as downstream AI task.} 
	\label{table1}
	\setlength{\abovecaptionskip}{-0.2cm}
	\begin{tabular}{lllllll}
		\toprule[1pt]
		Method & SI & ACC & F1-score & PSNR & MSE & SSIM \\
		\hline
		original & 987.53 & 85.45\% & 0.8544 & - & 0 & 1 \\
		TDIC & 53.87 & 63.26\% & 0.6353 & 23.49 & 0.0050 & 0.8100 \\
		APIC & 64.60 & 68.55\% & 0.6848 & 20.26 & 0.0104 & 0.7239 \\
		\hline
		SAIC & \textbf{68.10} & \textbf{69.69\%} & \textbf{0.6968} & 20.70 & 0.0094 & 0.7164 \\
		\toprule[1pt]
	\end{tabular}
\end{table}
\begin{table}
	\scriptsize
	\vspace{-0.4cm}
	\centering
	\caption{Classification results on the ImageNet dataset with 0.5bpp using ResNet18 as downstream AI task.}
	\label{table2}
	\setlength{\abovecaptionskip}{-0.2cm}
	\begin{tabular}{lllllll}
		\toprule[1pt]
		Method & SI & ACC & F1-score & PSNR & MSE & SSIM \\
		\hline
		original & 128.50 & 89.42\% & 0.8937 & - & 0 & 1 \\
		TDIC & 57.02 & 78.35\% & 0.7752 & 29.37 & 0.0016 & 0.8953 \\
		APIC & 80.60 & 84.53\% & 0.8451 & 26.44 & 0.0027 & 0.8637 \\
		\hline
		SAIC & \textbf{86.18} & \textbf{85.25\%} & \textbf{0.8522} & 26.40 & 0.0028 & 0.8391 \\
		\toprule[1pt]
	\end{tabular}
	\vspace{-0.5cm}
\end{table}
\begin{table}[t]
	\scriptsize
	\vspace{-0.3cm}
	\centering
	\caption{Classification results on the STL dataset with 0.125 bpp using VGG16 as downstream AI task.}
	\label{table3}
	\setlength{\abovecaptionskip}{-0.2cm}
	\begin{tabular}{lllllll}
		\toprule[1pt]
		Method & SI & ACC & F1-score & PSNR & MSE & SSIM \\
		\hline
		original & 672.02 & 88.94\% & 0.8899 & - & 0 & 1 \\
		TDIC & 42.29 & 57.04\% & 0.5733 & 23.55 & 0.0050 & 0.8042 \\
		APIC & 89.67 & 71.04\% & 0.7102 & 18.96 & 0.0137 & 0.6959 \\
		\hline
		SAIC & \textbf{91.65} & \textbf{71.99\%} & \textbf{0.7186} & 18.94 & 0.0138 & 0.6939\\
		\toprule[1pt]
	\end{tabular}
\end{table}
\vspace{-2.2ex}
\subsection{Parameter Analysis}
\vspace{-1ex}
\noindent\textbf{Ablation Experiments.} \;To evaluate the influence of GSW mechanism, we conduct ablation experiments. GSW get semantic weights of feature maps, which are used in ${\rm L_{SAIC}}$ to retain more semantic-level information. We conduct ablation experiment when feature maps have the same importance degree and no semantic difference. In that case, semantic weights equal to 1 for all feature maps. Thus, SAIC degenerates into APIC. APIC is the special case of SAIC when semantic-level of feature maps are equally important. That is to say, APIC is SAIC without GSW.
\begin{figure}
	\setlength{\abovecaptionskip}{-10pt}   
	\setlength{\belowcaptionskip}{-10cm}   
	\centering
	\includegraphics[width=0.7\linewidth]{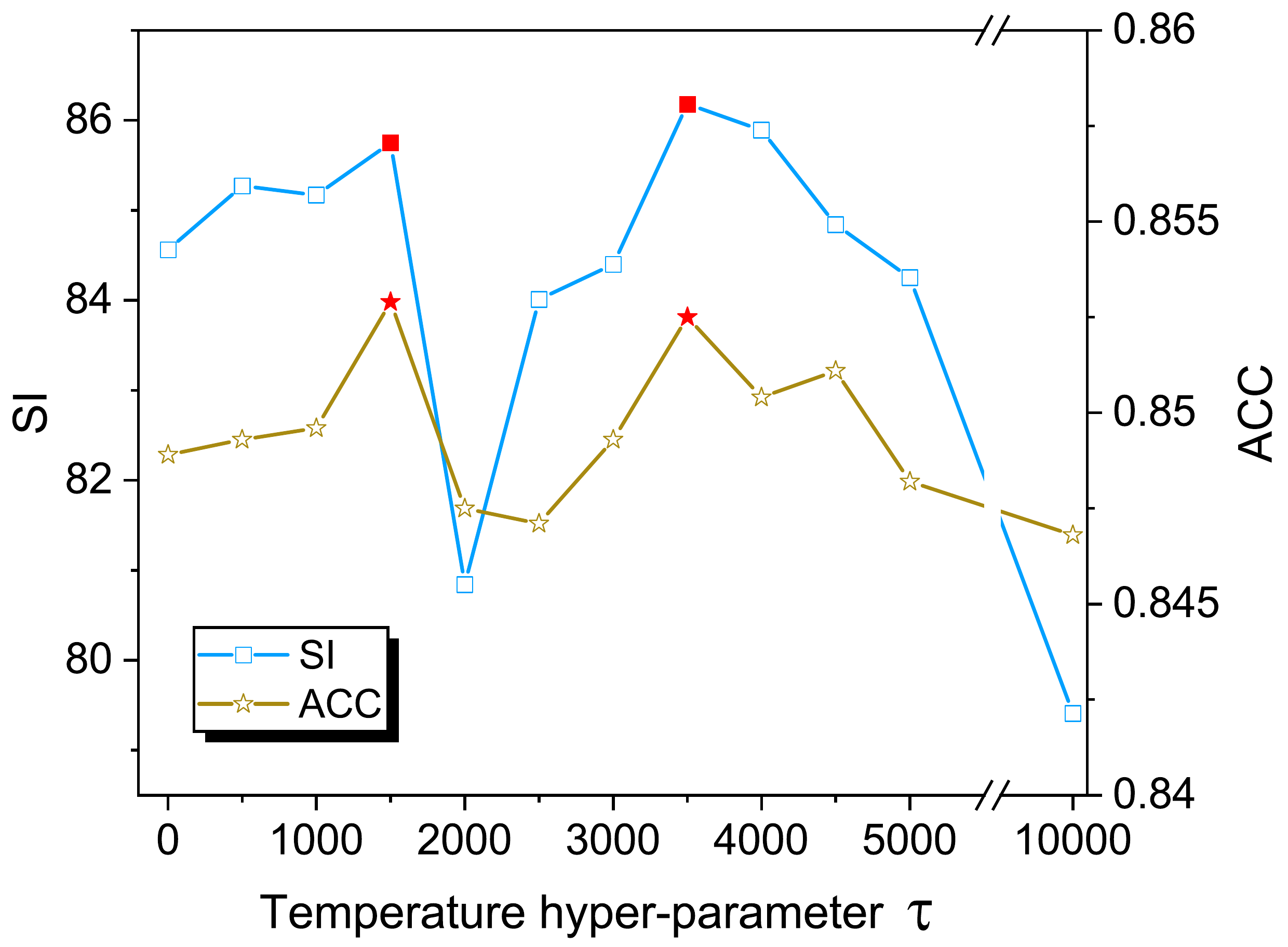}
	\vspace{-0.3cm}
	\caption{Classification results on ImageNet dataset with 0.5 bpp using different hyper-parameter $\tau$.}
	\label{tao}
	\vspace{-0.5cm}
\end{figure}

\noindent\textbf{Impact of Hyper-parameter.} \;To evaluate the influence of temperature hyper-parameter $\tau$, Fig.\ref{tao} shows the performance of SAIC with different temperature hyper-parameters $\tau$. It is significant in Fig.\ref{tao} that with $\tau$ increases, the overall variation trend of SI and ACC is first up, then down, then up and down again. This is because the hyper-parameter $\tau$ can control the degree of dispersion of semantic weights distribution, and extreme concentrated ($\tau=1, \tau=10000$) or decentralized ($\tau=2000$) semantic weights can degrade the performance. Excessively concentrated semantic weights almost the same importance degrees in different channels, and SAIC is close to APIC. Excessively decentralized semantic weights are equivalent to deleting some feature maps. Only appropriate degree of dispersion can have better performance improvement. The optimal value of $\tau$ varies with the model structure and the dataset. See appendix for semantic weights distribution diagram.
\begin{table}
	\scriptsize
	\vspace{-0.6cm}
	\centering
	\caption{Object detection results using RFBNet on the Pascal VOC dataset with 0.5 bpp.}
	\label{object detection}
	\setlength{\abovecaptionskip}{-0.2cm}
	\begin{tabular}{lllllll}
		\toprule[1pt]
		Method & mAP & PSNR & MSE & SSIM \\
		\hline
		original &  80.6\% & - & 0 & 1 \\
		TDIC & 71.3\% & 25.08  & 0.0040 & 0.7204 \\
		APIC & 72.1\% & 23.54 & 0.0054 & 0.6962 \\
		\hline
		SAIC & 72.0\% & 22.92 & 0.0060 & 0.6990 \\
		\toprule[1pt]
	\end{tabular}
	\vspace{-0.5cm}
\end{table}
\begin{table}
	\scriptsize
	\vspace{0.1cm}
	\centering
	\caption{Classification results on STL dataset with different compression ratios using ResNet18 as downstream AI task.}
	\label{table5}
	\begin{tabular}{llll}
		\toprule[1pt]
		ACC / SI & 0.125 bpp($\tau$ =400) & 0.25 bpp($\tau$ =3100) & 0.5 bpp($\tau$ =2900) \\
		\hline
		original & 85.45\% / 987.53	& 85.45\% / 987.53 & 85.45\% / 987.53 \\
		TDIC & 63.26\% / 53.87 & 76.19\% / 86.46 & 77.75\% / 94.52 \\
		APIC & 68.55\% / 64.60 & 77.66\% / 99.55 & 79.04\% / 106.86 \\
		\hline
		SAIC & 69.69\% / 68.10 & 78.43\% / 100.75 &	79.29\% / 106.44 \\
		\toprule[1pt]
	\end{tabular}
	\vspace{-0.5cm}
\end{table}
\vspace{-2ex}
\subsection{Experiments on Object Detection Task}
\vspace{-0.8ex}
The SAIC method has generalization ability, and the images compressed by SAIC can be used for a variety of downstream AI tasks. To verify generalization performance, the images compressed by SAIC can be applied to the object detection tasks. We select RFBNet \cite{liu2018receptive} trained on VOC0712 to conduct the object detection task. Table \ref{object detection} shows the mAP values of original and compressed images (bpp=0.5) using different methods on VOC 2007 validation. As shown in Table \ref{object detection}, the task performance (mAP) of APIC and SAIC is comparable and higher than TDIC. To further improve task performance on object detection, we can utilize semantic-level information of object detection task to train the compression network in the future.
\vspace{-0.3cm}
\subsection{Compression Ratio}
\vspace{-1ex}
Traditional image compression uses rate-distortion to evaluate compression performance. Compression ratio represents compression degree and pixel-level distortion (e.g.,PSNR, MSE) represents distortion degree. We use semantic-level rate-distortion to evaluate compression performance. Compression ratio represents compression degree while SI and ACC represent semantic-level distortion. Table \ref{table5} shows SI and ACC of TDIC, APIC and SAIC methods with 0.125 bpp, 0.25 bpp and 0.5 bpp. Thanks to the quantizer, accurate compression ratios can be calculated. Different compression ratios can be obtained by controlling the size of quantized feature maps ${\boldsymbol{q}}$. As shown in Table \ref{table5}, at the same compression ratio, SAIC can get less semantic-level distortion. In turn, at the same semantic-level distortion, SAIC can compress more. As the compression ratio shrinks, the performance improvement getting higher and higher. That's because with the degree of compression decreasing, the compressed image is less distorted and leaves less room for improvement. Some images can be correctly perceived without the assistance of semantic-level information.
\vspace{-0.4cm}
\section{Conclusion}
\vspace{-0.1cm}
This work proposed a novel SAIC method, which takes downstream AI tasks into consideration. In particular, the proposed SAIC innovatively aims at maximizing the semantic-level information required by downstream AI tasks during compression process, so as to improve the downstream AI tasks' performance. A new metric SI has also been proposed to quantify the semantic-level distortion during image compression. Experimental results show that SAIC can achieve 10.16\% and 1.66\% higher ACC values on STL dataset with 0.125 bpp than TDIC and APIC, respectively. The proposed SAIC method can take into account human visual experience and machine perception performance, and can be used for a variety of intelligent tasks. In the future, it can be deployed for other applications such as denoising and super-resolution.
\vspace{-0.3cm}
\bibliographystyle{IEEEbib}
\bibliography{icme2022template}

\end{document}